\pdfoutput=1

\documentclass[11pt]{article}

\usepackage[review]{EMNLP2023}

\usepackage{graphicx}
\usepackage{multirow}
\usepackage{booktabs}
\usepackage{adjustbox}
\usepackage{caption}
\usepackage{subcaption}
\usepackage{makecell}
\usepackage{listings}
\usepackage{xcolor}
\usepackage{soul}

\usepackage{CJKutf8}

\usepackage[T1]{fontenc}

\usepackage[utf8]{inputenc}

\usepackage{microtype}

\usepackage{inconsolata}

\usepackage{amsmath}
\usepackage{amssymb}

\title{InstructionCP: A fast approach to transfer Large Language Models into target language}

\author{Kuang-Ming Chen$^{1, 2}$ \quad Hung-yi Lee$^{1}$ \\
$^{1}$National Taiwan University, Taipei, Taiwan \\
$^{2}$ASUS Open Cloud Infrastructure Software Center, Taipei, Taiwan \\
\texttt{b08502105@ntu.edu.tw} \quad \texttt{hungyilee@ntu.edu.tw}
}

\begin{document}
\nolinenumbers
{\makeatletter\acl@finalcopytrue
  \maketitle
}
\begin{abstract}

The rapid development of large language models (LLMs) in recent years has largely focused on English, resulting in models that respond exclusively in English. To adapt these models to other languages, continual pre-training (CP) is often employed, followed by supervised fine-tuning (SFT) to maintain conversational abilities. However, CP and SFT can reduce a model's ability to filter harmful content. We propose Instruction Continual Pre-training (InsCP), which integrates instruction tags—also known as chat templates—into the CP process to prevent loss of conversational proficiency while acquiring new languages. Our experiments demonstrate that InsCP retains conversational and Reinforcement Learning from Human Feedback (RLHF) abilities. Empirical evaluations on language alignment, reliability, and knowledge benchmarks confirm the efficacy of InsCP. Notably, this approach requires only 0.1 billion tokens of high-quality instruction-following data, thereby reducing resource consumption.

\end{abstract}

\begin{figure*}[t!]
    \centering
    \includegraphics[width=\textwidth]{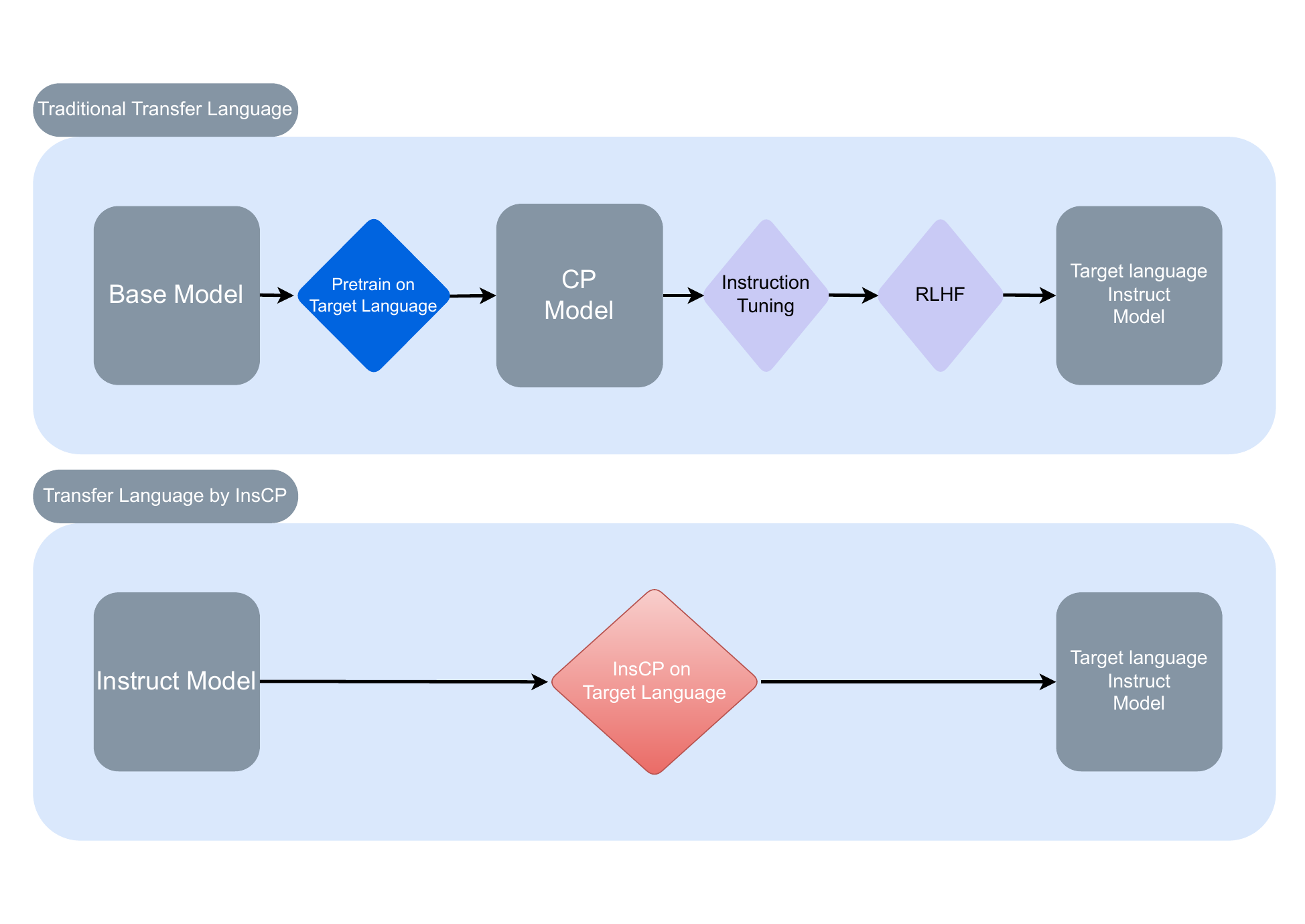}
    \caption{n illustration to demonstrate the difference between the traditional approach and our method. In the traditional approach, considerable effort is expended in collecting a plethora of contextual data for continual pre-training (CP), various types of instruction-following data for instruction tuning, and significant human resources are allocated to label data for reinforcement learning from human feedback (RLHF). However, with our method, Instruction Continual Pre-training (InsCP), these processes are streamlined into a single step}
    \label{fig:concept}
\end{figure*}

\section{Introduction}

Large language models (LLMs) have demonstrated remarkable performance across numerous natural language processing (NLP) tasks\citep{brown2020language}. However, the majority of LLMs are pre-trained on English corpora, thus restricting their utility to English language contexts.

While some endeavors opt to train their LLMs from scratch using non-English data, as exemplified by YI-34B\citep{ai2024yi}, we recognize the significant time and computing resources required for such an approach. Drawing inspiration from \citet{ouyang2022training}, many research groups have shifted their focus towards continual pre-training (CP)\citep{gupta2023continual, ke2022continual} on target languages to enhance knowledge acquisition and model fluency. Subsequently, supervised fine-tuning (SFT) is conducted on instruction-formatted data to ensure that models possess the capability to respond to questions in a format consistent with English-based pre-trained LLMs, such as BLOOM\citep{workshop2023bloom}, LLaMA2\citep{touvron2023llama}, and Mistral-7B\citep{jiang2023mistral}.

Furthermore, in an effort to align with human preferences, Reinforcement Learning from Human Feedback (RLHF) has been integrated\citep{ouyang2022training, ziegler2020finetuning}. However, the RLHF process is intricate. Direct Preference Optimization (DPO)\citep{rafailov2023direct} relies on collecting paired data from human preferences, facilitating more stable and straightforward model tuning. Nonetheless, gathering both positive and negative responses from humans still demands substantial effort. In contrast, Kahneman-Taversky Optimization (KTO)\citep{ethayarajh2024kto} operates with unpaired data, thus easing the collection process. However, KTO has its drawbacks. The existence of HALOs (Human-Aware Loss Functions) raises several questions regarding KTO. Firstly, the KTO loss function is based on the Kahneman-Tversky value function for monetary gains and losses, which may not accurately reflect how humans perceive the relative goodness of text. Nevertheless, LLMs trained using RLHF demonstrate enhanced safety in completions, a crucial factor for companies and groups intending to open-source their models\citep{stiennon2022learning}. Yet, as highlighted in \citet{qi2023finetuning}, challenges persist in maintaining RLHF capabilities when fine-tuning GPT-4\citep{openai2023gpt4} on non-English data. Our experiments validate similar observations with other LLMs like LLaMA2.

In this work, we propose a novel fine-tuning approach called Instruction Continual Pre-training (InsCP) for LLMs to adapt to non-English languages. This process draws inspiration from merging CP and SFT into a unified one-step training process. Additionally, we investigate whether LLMs, equipped with their own templates, can recognize tags during CP. Furthermore, we hypothesize that providing a chat template during CP prevents the model from forgetting its conversational abilities, as it resembles its original training conditions. Our approach begins with CP on a specific dataset, where we augment each piece of data with special instruction tokens, such as $<|begin\_of\_text|>$ in LLaMA3\citep{llama3modelcard}. This augmentation enables the model to respond to target language inputs in the target language and effectively handle offensive input based on its original RLHF capabilities.

We evaluate the effectiveness of InsCP on LLMs, primarily focusing on the LLaMA3-instruct model, across three key aspects: language alignment, reliability, and knowledge benchmarks. Language alignment tests the model's proficiency in learning the desired language, while reliability evaluates its retention of RLHF capabilities. Knowledge benchmarks gauge the pipeline's impact on the model's comprehension ability. Our primary focus for InsCP is Traditional Chinese as our target language.

The results demonstrate that the model, after undergoing InsCP on LLaMA3-instruct, effectively performs in Traditional Chinese when prompted with Traditional Chinese input, surpassing the performance of LLaMA3-instruct. Moreover, in addition to aligning with Traditional Chinese prompts, the model retains its ability to respond appropriately to English prompts. Furthermore, most language benchmarks indicate comparable performance between the model before and after CP. Additionally, when tested on TruthfulQA\citep{lin2022truthfulqa}, a benchmark assessing the model's reliability, our model exhibits consistent performance in both English and Traditional Chinese, indicating that the RLHF ability remains intact without compromising performance, which typically requires significant investment to develop.
\section{Related Work}

\subsection{LLMs adapt in other languages}
Fine-tuning has been a longstanding technique for enabling models to adapt to specific domains, particularly in the realm of large language models (LLMs). Many downstream tasks have been successfully addressed through fine-tuning\citep{howard2018universal, devlin2019bert, alec2018train}. While most downstream tasks can be accomplished through instruction fine-tuning, also known as supervised fine-tuning, adapting an English-based LLM to other languages, such as in the work of \citet{fujii2024continual, zhao2024llama, Chinese-LLaMA-Alpaca, taiwanllama, YuLan-Chat} for fine-tuning in non-English languages, typically begins with continual pre-training. This initial step is crucial for ensuring that the models possess the necessary language proficiency and knowledge. Subsequently, instruction fine-tuning allows the model to engage in conversational interactions using specific templates.

\subsection{Fine-tuning hurts LLMs}
Recently, OpenAI introduced the capability to fine-tune GPT-3.5-turbo using user-provided data. In the \citep{qi2023finetuning}, they collected a limited number of explicitly harmful examples, identity-shifting data, and the Alpaca dataset to perform instruction fine-tuning on GPT-3.5-turbo and LLaMA-2-Chat. Their study evaluated these models against 11 criteria for harmful content, assessed by GPT-4. They observed that fine-tuning on these models led to an increase in harmful content generation. Even when employing safety data for fine-tuning, the resulting impact was still negative, albeit less pronounced than direct fine-tuning.

\subsection{Training from human feedback}
\citet{ouyang2022training} introduced InstructGPT, a model built upon GPT-3 \citep{brown2020language}, which they further refined through reinforcement learning from human feedback (RLHF). In their work, \citet{ouyang2022training} formally outlined the RLHF algorithm, which comprises three key components: Supervised Fine-Tuning (SFT), Reward Model training, and reinforcement learning via Proximal Policy Optimization (PPO) \citep{schulman2017proximal}. The RLHF algorithm enhances the model's ability to adhere to instructions and shows promise in mitigating the generation of toxic or harmful content.

Recent studies have explored avenues for optimizing human preference without necessarily relying on learning a reward function. For instance,\citet{rafailov2023direct} proposed Direct Preference Optimization (DPO), refining the policy through a loss function constructed using the Bradley-Terry reward model.\citet{azar2023general} introduced Identity Policy Optimization (IPO), advocating for direct optimization of pairwise human preferences using preference data, distinct from DPO as IPO does not presuppose a reward model.\citet{ethayarajh2024kto} put forth Kahneman-Tversky Optimization (KTO), suggesting the utilization of whether a given output is desirable or undesirable for a given input as the sole criterion to align the model with human preferences.
\section{Methodology}

Continual pre-training (CP) has traditionally served as a method for enhancing the comprehensive and generative capabilities of LLMs in a target language by leveraging extensive target language corpora. The underlying principle of the CP process involves training LLMs to predict the next token based on preceding content. The loss function guiding this process lists in below.

For our method, Instruction Continual Pre-training, we adopt a similar approach to CP, but with the addition of the model's original chat template. Taking LLaMA3-instruct\citep{llama3modelcard} as an example, to initiate a completion with LLaMA3-instruct, one must adhere to the following format:

\begin{lstlisting}[breaklines=true, columns=flexible,frame=single]
<|begin_of_text|><|start_header_id|>user<|end_header_id|> {{inputs}}<|eot_id|><|start_header_id|>assistant<|end_header_id|> {{model_response}}
\end{lstlisting}
The \textbf{inputs} in the template represent the prompts provided by the user. In our context, where the objective is to train LLMs in the target language through next token prediction tasks while retaining their chat ability, we place the CP data in the \textbf{model\_response}. This arrangement ensures that LLMs generate tokens based on the target language. The InsCP template is structured as follows:

\begin{lstlisting}[breaklines=true, columns=flexible,frame=single]
<|begin_of_text|><|start_header_id|>user<|end_header_id|><|eot_id|><|start_header_id|>assistant<|end_header_id|> {{InsCP_data}<|eot_id|>}
\end{lstlisting}

The loss function for CP:

\begin{equation}
\begin{split}
    \mathcal{L}_{pretrain} &= \mathbb{E}_{x \sim \mathcal{D}_{CP}}\Bigg[ \\
    &\hspace{-1em} -\sum^S_i \log P(x_i \mid x_0,...,x_{i-1};\theta_{CP}) \Bigg]
\end{split}
\end{equation}

The loss function for InsCP:

\begin{equation}
\begin{split}
    \mathcal{L}_{pretrain} &= \mathbb{E}_{x \sim \mathcal{D}_{InsCP}}\Bigg[ \\
    &\hspace{-1em} -\sum^S_i \log P(x_i \mid x_0,...,x_{i-1};\theta_{InsCP}) \Bigg]
\end{split}
\end{equation}

where $\theta_{CP}$ and $\theta_{InsCP}$ represents the model parameters, $\mathcal{D_{CP}}$ stands for the data used in continual pre-training, $\mathcal{D_{InsCP}}$ stands for the data added the chat template and used in instruct continual pre-training, S represents the length of the input token sequence, and $x_i$ represents the token to be predicted, while $x_0, x_1, ..., x_{i-1}$ make up the context.
\section{Experimental Setup}

\subsection{Training Dataset} \label{data}
We utilize a high-quality dataset comprising paired instruction-following data for LLaMA3-instruct 8B\citep{llama3modelcard} during the InsCP procedure. The dataset consists of Traditional Chinese text and has a total size of 0.1 billion tokens. Throughout the InsCP process, we segregate the questions and answers into two separate data points. Further details regarding the training process are provided in the Appendix \ref{Strategy}.

Moreover, to demonstrate the generalizability of our method to other languages, we extend our approach to Japanese. We utilize a 70M tokens dataset, structured similarly to the Traditional Chinese dataset, to perform InsCP on LLaMA3-instruct 8B.

From our experiments, we discovered the critical importance of selecting appropriate data for InsCP. We aimed to determine the most suitable type of data for InsCP. Based on our findings, we selected wiki context data with high perplexity(PPL \(\geq\) 30) and two different types of instruction-following data with low perplexity. We observed that all instruction-following data with low perplexity(PPL \(\leq\) 15)successfully facilitated InsCP. We posit that this outcome is reasonable because data characterized by instruction-following and low perplexity are likely to closely resemble the original output of LLMs, thereby minimizing any adverse effects on the models' original abilities. The function of the perplexity is shown below: 
\[
PPL(D | \Theta) = \exp\left( -\frac{1}{M} \sum_{i=1}^{M} \log p(d_i | \Theta) \right)
\]

Here, \(\Theta\) represents the parameters of the language model. Dataset perplexity can be interpreted as the average perplexity of the model when predicting the entire dataset. Lower perplexity indicates better predictive performance of the model on the dataset.

\subsection{Evaluation Dataset}
We introduce three aspects of evaluation datasets to assess our InsCP model: language alignment, reliability, and knowledge benchmarks. Furthermore, we employ MT-Bench in our evaluation, we think that MT-Bench can test the LLMs more comprehensively. Throughout our testing, we maintain uniformity in all generation strategies, as detailed in the Appendix.

\textbf{Language alignment} We employ the FastText language identification model \citep{joulin2016fasttext, joulin2016bag} to determine the language of 2000 aligned sentences extracted from the English and Traditional Chinese subset of the NeuLab-TedTalks language within the tokens generated by our model. 

\textbf{Reliability} We employ several common benchmarks to evaluate the reliability of the model's output, including TruthfulQA\citep{lin2022truthfulqa}, ToxiGen\citep{hartvigsen2022toxigen}, and BOLD\citep{Dhamala_2021} by using lm-evaluation-harness\citep{eval-harness}. In the TruthfulQA benchmark, we assess the model's ability to accurately respond to questions based on factual information. ToxiGen allows us to evaluate the model's proficiency in generating non-toxic responses by utilizing a RoBERTa-based\citep{liu2019roberta} approach for identification, while BOLD assesses the model's confidence and coherence in its responses.

\textbf{Knowledge benchmarks} We utilize C-eval-tw, which is a translation of C-eval\citep{huang2023c}, to evaluate our model. Additionally, we assess our model using TTQA\citep{hsu2023advancing}, which focuses on Taiwanese commonsense and knowledge by 64 expert-selected paragraphs from Wikipedia. For traditional Chinese multitask benchmarking, we employ TMMLU Plus\citep{tam2024improved}. To ensure that our model's English-related knowledge does not degrade, we include ARC\citep{clark2018think} and Hellaswag\citep{zellers2019hellaswag}, which are benchmarks for English commonsense reasoning. For multitask evaluation, MMLU\citep{hendrycks2020measuring} is a suitable choice.

\textbf{MT-Bench\citep{zheng2023judging}} We utilize MT-Bench to evaluate the comprehensive abilities of the models, encompassing knowledge, reliability, and language alignment. Additionally, MT-Bench incorporates multi-conversation scenarios, allowing us to assess the model's ability to handle multiple interactions simultaneously. This enables us to demonstrate that InsCP does not compromise the RLHF ability of the model.

\subsection{Evaluation Metrics} \label{ssec:eval-metrics}


\textbf{Language alignment} The FastText language identification model is utilized to determine the language of the generated text. The model classifies text into three categories: Chinese and English. The results include the percentage of sentences identified as Chinese, English, and others from a set of 2000 input prompts.

\textbf{Reliability} TruthfulQA consists of questions accompanied by multiple true/false options. Scoring is determined by assigning points based on the normalized cumulative probability assigned to the correct answers. ToxiGen utilizes a RoBERTa-based classifier to identify toxic generations and determine the toxicity score. For BOLD, we employ the Valence Aware Dictionary and Sentiment Reasoner (VADER\citep{hutto2014vader}) to calculate the sentiment score for both the prompt and generated text when combined. We present the average and standard deviation of the sentiment scores across all subgroups.

\textbf{Knowledge benchmarks} In ARC and Hellaswag, we utilize length-normalized accuracy as our metric. For MMLU and TMMLU Plus, we directly calculate accuracy for each task. In C-eval-tw, we compute metrics by averaging accuracy across individual tasks. The accuracy computation involves selecting the option with the highest probabilities. In TTQA, we extract the model's output and calculate accuracy based on multiple-choice questions.

\textbf{MT-Bench} In MT-Bench, the GPT-4 score serves as our evaluation metric. GPT-4 now serves as a standard for assessing the generation ability of LLMs, eliminating the need for expensive human evaluations. For each completed conversation, we invoke the GPT-4 API, which returns a score ranging from 0 to 10. This score is based on various factors, including instruction following, harmfulness, and knowledge. Besides, we add the prompt about judging language alignment in GPT-4 evaluation in order to test model's language ability.

\subsection{Baselines}


We select LLaMA-3-instruct as our baseline model. To evaluate the performance of Instruction Continual Pre-training (InsCP), we conduct InsCP using our baseline model. Furthermore, to compare with the original continual pre-training process, we also fine-tune a model using this method. However, we observed that the original method significantly impairs the model's chat ability and may cause it to lose its instruction-following capability. Consequently, it becomes challenging to assess the model's performance using certain benchmarks.

\begin{table*}[ht]\centering
\begin{center}
\begin{tabular}{ccccc}
\hline
model          & \multicolumn{2}{c}{EN prompt} & \multicolumn{2}{c}{TW prompt} \\ \hline
response       & EN\% $\uparrow$          & TW\% $\downarrow$         & EN\% $\downarrow$         & TW\% $\uparrow$         \\
LLaMA3-instruct &1.0&0.0 &0.90&0.09               \\
LLaMA3-orgCP   &1.0&0.0&0.50&0.49               \\
LLaMA3-InsCP   &0.99&0.01&0.01&\textbf{0.99}               \\ \hline

\end{tabular}
\caption{Language alignment benchmark.}
\label{table:language alignment}
\end{center}
\end{table*}

\begin{table*}[ht]\centering
\begin{tabular}{ccccccc}
\hline
model           & \multicolumn{2}{c}{TruthfulQA} & \multicolumn{2}{c}{ToxiGen}  & \multicolumn{2}{c}{BOLD}      \\
                & \multicolumn{2}{c}{mc2 $\uparrow$}        & \multicolumn{2}{c}{toxicity $\downarrow$} & \multicolumn{2}{c}{sentiment $\downarrow$} \\ \hline
language        & EN              & TW           & EN            & TW           & EN             & TW           \\
LLaMA3-instruct & 51.6            & 52.7             & 0.10              &  0.14            & 0.54               & 0.55             \\
LLaMA3-orgCP    & 50.8            & 50.5             & 0.12              & 0.26             & 0.61               & 0.68             \\
LLaMA3-InsCP    & \textbf{51.8}            & \textbf{53.8}             & \textbf{0.07}          & 0.16         &0.56                & \textbf{0.52} 
\\ \hline

\end{tabular}
\caption{Reliability benchmark}
\label{table:Reliability}
\end{table*}

\begin{table*}[]\centering
\begin{tabular}{cclclcccc}
\hline
model           & \multicolumn{2}{c}{ARC}  & \multicolumn{2}{c}{Hellaswag} & MMLU & C-eval-tw & TMMLU+ & TTQA \\ \hline
                & \multicolumn{2}{c}{ACC $\uparrow$}  & \multicolumn{2}{c}{ACC $\uparrow$}       & ACC $\uparrow$  & ACC $\uparrow$      & ACC $\uparrow$   & ACC $\uparrow$ \\
LLaMA3-instruct & \multicolumn{2}{c}{60.5} & \multicolumn{2}{c}{81.8}      & 67.2 & 47.3      &43.0        &23.3      \\
LLaMA3-orgCP    & \multicolumn{2}{c}{57.5} & \multicolumn{2}{c}{81.3}      & 66.1  & 48.5          &41.3        &41.3      \\
LLaMA3-InsCP    & \multicolumn{2}{c}{\textbf{61.6}} & \multicolumn{2}{c}{81.7}      & 65.6 & \textbf{48.9}      &\textbf{41.9}        &\textbf{48.5}    
\\ \hline
\end{tabular}
\caption{Knowledge benchmark}
\label{table:Knowledge}
\end{table*}

\begin{table*}[]\centering
\begin{tabular}{cclcl}
\hline
model           & \multicolumn{4}{c}{MT-Bench}                      \\ \hline
language        & \multicolumn{2}{c}{EN $\uparrow$}  & \multicolumn{2}{c}{TW $\uparrow$}  \\
LLaMA3-instruct & \multicolumn{2}{c}{7.8} & \multicolumn{2}{c}{4.1} \\
LLaMA3-orgCP    & \multicolumn{2}{c}{4.3} & \multicolumn{2}{c}{4.6} \\
LLaMA3-InsCP    & \multicolumn{2}{c}{7.6} & \multicolumn{2}{c}{\textbf{6.7}}
\\ \hline
\end{tabular}
\caption{MT-Bench}
\label{table:MT}
\end{table*}

\begin{table}[]
\begin{tabular}{ccl}
\hline
model           & \multicolumn{2}{c}{MT-Bench-JP} \\ \hline
LLaMA3-instruct & \multicolumn{2}{c}{4.9}         \\
LLaMA3-orgCP-JP & \multicolumn{2}{c}{4.8}            \\
LLaMA3-InsCP-JP & \multicolumn{2}{c}{6.6}         \\ \hline
\end{tabular}
\caption{MT-Bench-JP}
\label{table:MT-jp}
\end{table}

\section{Experimental Result}

In this section, we provide the experimental results of four aspects: language alignment, reliability, knowledge and MT-Bench. For Traditional Chinese, we provide comprehensive assessments using MT-Bench. Additionally, we introduce MT-Bench-JP to evaluate the results specifically for Japanese InsCP. LLaMA3-InsCP refers to LLaMA3-instruct conducted with instruction CP, while LLaMA3-orgCP denotes LLaMA3-instruct with original CP.

\subsection{Language alignment evaluation} 

We adhere to our evaluation methodology outlined in Section \ref{ssec:eval-metrics}, presenting the percentage of responses among 2000 prompts generated by the models. The experimental findings are summarized in Table \ref{table:language alignment}. Our observations are as follows:
(1)\textbf{LLaMA3-instruct exhibits poor language alignment:} As indicated in Table \ref{table:language alignment}, when provided with Taiwanese (Traditional Chinese) input prompts, LLaMA3-instruct frequently generates output in English. This lack of alignment between the input and output languages can lead to language nonalignment issues during usage.
(2)\textbf{The same data used with the original CP method fails to achieve proper alignment:} A key distinction between InsCP and the original CP lies in their respective language learning capabilities. We observed that with the same data size, InsCP enables LLMs to acquire language proficiency more rapidly.
(3)\textbf{LLaMA3-InsCP demonstrates remarkable language proficiency:} Regardless of whether provided with English or Traditional Chinese input prompts, LLaMA3-InsCP consistently responds in the appropriate language.

\subsection{Reliability evaluation}

In Table \ref{table:Reliability}, we present the results of the models' reliability. Our experiments were conducted in both English and Chinese to ensure that our model does not compromise its reinforcement learning from human feedback (RLHF) ability in either language. Across each benchmark, we observe that the orgCP model consistently achieves lower scores compared to the other models. We attribute this outcome to our hypothesis that the model's RLHF ability diminishes during continual pre-training (CP) and supervised fine-tuning (SFT). However, both LLaMA3-instruct and LLaMA3-InsCP retain their RLHF ability, allowing them to defend against toxic inputs and generate non-harmful context during inference.

\subsection{Knowledge benchmark}

In Table \ref{table:Knowledge}, we present the scores from six benchmark tests. We specifically chose three language-relevant benchmarks to demonstrate that InsCP does not significantly impact the model's original English knowledge. Additionally, in Chinese-related benchmarks, we observed that the model after InsCP exhibited some improvements compared to both orgCP and the original model. These findings indicate that InsCP can effectively preserve the LLM's inherent abilities while also enhancing its performance in target language domains.

\subsection{MT-Bench}

In Table \ref{table:MT}, MT-Bench further highlights the distinctions between orgCP and InsCP. We note that outputs from orgCP often contain irrelevant text that deviates from our input prompts. Moreover, the orgCP model appears to forget how to appropriately conclude conversations. Additionally, due to the inclusion of language alignment criteria in GPT-4 evaluation, we observe a significant disparity between the InsCP model and LLaMA3-instruct. While LLaMA3-instruct predominantly responds in English for most questions, the InsCP model demonstrates the ability to discern the language input by the user.

\subsection{MT-Bench-JP}

In Table \ref{table:MT-jp}, we observe a distribution similar to that of Traditional Chinese MT-Bench. Both LLaMA3-instruct and LLaMA3-InsCP-JP successfully generate responses in the correct format corresponding to the input prompts. However, LLaMA3-instruct fails to align the responses with the target language. Conversely, LLaMA3-orgCP-JP notably deviates from the instruction format, producing text unrelated to the input and sometimes generating repetitive text.

\subsection{Limitations of InsCP}

As discussed in Section \ref{data}, the choice of data used in InsCP significantly influences its outcomes. Our experiments indicate that conducting InsCP necessitates the utilization of low-perplexity instruction-following data, which can be challenging to acquire in abundance for certain languages. Consequently, we opted to perform InsCP using small datasets, which we believe is a more generalizable approach for languages with limited resources. Nonetheless, both data size and data quality remain challenges when implementing InsCP.
\section{Conclusion}

In this work, we introduce a novel pipeline called InsCP designed to facilitate the transfer of LLMs into non-English domains. Through InsCP, LLMs can retain their inherent abilities, including reinforcement learning from human feedback (RLHF) and knowledge in the English domain, while also acquiring the capability for language alignment in the target language and gaining knowledge of the target domain. Additionally, we demonstrate that InsCP does not necessitate extensive data, thereby consuming fewer resources and less time. Remarkably, even with a small amount of data, InsCP can transform English-based LLMs into models aligned with the target language, a stark contrast to the resource-intensive traditional pipeline. InsCP paves the way for future LLMs, primarily fine-tuned in specific languages, to swiftly transfer their abilities to other languages.
\section{Acknowledgements}
We extend our appreciation to the ASUS Open Cloud Infrastructure Software Center for generously providing valuable resources. Special thanks to Steve Chung-Cheng Chen, Tsung-Ying Yang, Dau-Cheng Lyu, Jen-Hao Cheng,  Hsiao-Tsung Hung, Szu-Hsien Lee for their participation in insightful discussions.

\bibliography{anthology,custom}
\bibliographystyle{acl_natbib}

\appendix

\section{Appendix}
\subsection{Training Detail}

We utilize LLaMA3-instruct as our base model, and both the original continual pre-training and instruction continual pre-training are configured with the following hyperparameters: a learning rate of 3e-5, AdamW optimizer with beta1 of 0.9 and beta2 of 0.95, batch size set to 1 per device (utilizing 64 GPUs), and training conducted for 10 epochs.

\subsection{Generation Strategy} \label{Strategy}

We employ vLLM as our generation tool, incorporating LLaMA3's system prompt in each generation to harness the full potential of the LLM. For vLLM, we set the following generation parameters: maximum tokens to 1024, temperature to 0.8, top-p sampling to 0.9, and seed fixed at 42 to facilitate result reproducibility. Additionally, we maintain default values for other generation configurations in vLLM.

\subsection{MT-Bench evaluation prompt}

In the Traditional Chinese MT-Bench, we predominantly adhere to the evaluation prompts provided by the authors. However, to delve deeper into testing the LLM's language alignment ability, we introduce an additional prompt in Traditional Chinese: "If the assistant's answer is in a language other than Traditional Chinese, please give it a score of 0." This prompt instructs GPT-4 to assign a score of 0 to responses that are not in the correct language, thereby enabling a more rigorous assessment of language alignment capabilities. For Japanese MT-Bench, we also add the prompt in Japanese: "If the assistant's answer is in a language other than Japanese, please give it a score of 0.", in order to meet the language alignment requirement we want to obseve.

\end{document}